\documentclass[arxiv]{melba}

\usepackage{mwe} 

\usepackage{amsmath,amsfonts}
\setcitestyle{square}

\melbaid{YYYY:NNN}  
\doi{10.59275/j.melba.2025-AAAA}
\melbaauthors{Alex Michie and Simon J Doran}  
\email{Simon.Doran@icr.ac.uk}
\volume{2}
\firstpageno{1}  
\melbayear{2025}  
\datesubmitted{yyyy-m1-d1}  
\datepublished{yyyy-m2-d2}  

\melbaspecialissue{Medical Image De-Identification Benchmark (MIDI-B) 2025}
\melbaspecialissueeditors{Linmin Pei and Keyvan Farahani}

\ShortHeadings{Image deidentification in the XNAT ecosystem}{Alex Michie and Simon J Doran}

\title{Image deidentification in the XNAT ecosystem: use cases and solutions}

\author{
	\firstname Alex \surname Michie\aff{1}\orcid{0000-0003-1685-6092},
	\name Simon J Doran\aff{1}\orcid{0000-0001-8569-9188}
}
\affiliations{
	\num 1 \addr Division of Radiotherapy and Imaging, The Institute of Cancer Research, London
}

\abstract{
    XNAT is a server-based data management platform, with user-facing webapp, that is widely used in academia for curating large databases of DICOM images for research projects. In this contribution, we describe in detail a deidentification workflow for DICOM data using facilities in XNAT, together with independent tools in the XNAT ``ecosystem''. We list a number of different contexts in which deidentification might be needed, based on our prior experience. The starting point for participation in the Medical Image De-Identification Benchmark (MIDI-B) challenge was a set of pre-existing local methodologies, which were adapted during the validation phase of the challenge. Our result in the test phase was 97.91\%, considerably lower than our peers, due largely to an arcane technical incompatibility of our methodology with the challenge's Synapse platform, which prevented us receiving feedback during the latter parts of the validation phase. Post-submission, additional discrepancy reports from the organisers and via the MIDI-B Continuous Benchmarking facility, enabled us to improve this score significantly to 99.61\%. An entirely rule-based approach was shown to be capable of removing all name-related information in the test corpus, but exhibited failures in dealing fully with address data. Initial experiments using published machine-learning models to remove addresses were partially successful but showed the models to be ``over-aggressive'' on other types of free-text data, leading to a slight overall degradation in performance to 99.54\%. Future development will therefore focus on improving address-recognition capabilities, but also on better removal of identifiable data burned into the image pixels. Several technical aspects relating to the ``answer key''' are still under discussion with the challenge organisers, but we estimate that our percentage of ``genuine'' deidentification failures on the MIDI-B test corpus currently stands at 0.19\%.    
	}


\begin{document}

\twocolumn[\maketitle]

\section{Introduction}
    \subsection{Background and research context}
        XNAT\footnote{\url{https://www.xnat.org/}} is a server-based research platform for managing medical images that is particularly suitable for hosting canonical datasets for multicentre academic clinical studies. Users interact with XNAT via a sophisticated web application, and it provides flexible facilities for secure anonymisation, archiving, sharing, visualisation, annotation and processing of image data. In a UK context, widespread adoption of XNAT, with consequent expertise available at academic radiology centres, is an invaluable legacy from previous successful consortia funded by government and charities, viz.:
        \begin{itemize}
        \item the Cancer Research UK (CRUK) National Cancer Imaging Translational Accelerator (NCITA) \citep{mcateer2021introduction} and Radiation Research Network (RadNet)\footnote{\url{https://www.cancerresearchuk.org/funding-for-researchers/our-research-infrastructure/radnet-our-radiation-research-network}};
        \item the Dementias Platform UK (DPUK) \citep{bauermeister2020dementias} and UK Renal Imaging Network (UKRIN)\footnote{\url{https://www.kidneyresearchuk.org/research/research-networks/uk-renal-imaging-network/}}, supported by the UK's Medical Research Council;
        \item the FLIP programme\footnote{\url{https://www.aicentre.co.uk/our-platforms\#tab-2}} from the AI Centre for Value-Based Healthcare, supported by Innovate UK;
        \item the Radiotherapy Trials Quality Assurance Network (RTTQA)\footnote{\url{https://rttrialsqa.org.uk/}}, funded by the National Institute for Health and Care Research (NIHR).
        \end{itemize}
    
        \noindent Internationally, XNAT powers the EU’s Horizon2020 EuCanImage programme (incorporating data for both the HealthRI and Euro-BioImaging consortia) and the Australian Imaging Service, with large XNAT installations at major cancer centres throughout Europe, North America (e.g, \citep{gurney2017washington}) and South-East Asia.

        In our setting, The Institute of Cancer Research (ICR) hosts several XNAT instances to handle data stored under different information governance (IG) regimes. Inside the hospital network of our partner organisation, the Royal Marsden Hospital (RM), a secure XNAT server receives identifiable research data directly from the hospital's picture archiving and communications system (PACS) or vendor neutral archive (VNA). An in-house Java application, incorporating the DicomEdit 6 library\footnote{\url{https://wiki.xnat.org/xnat-tools/dicomedit}}, retrieves custom DicomEdit anonymisation scripts from a destination XNAT server, located either within the hospital’s Trusted Research Environment (TRE) or in the ICR. The application communicates with the source (“identifiable”) XNAT server, performs the relevant deidentification on RM hardware and transmits the results to the destination (``deidentified'') XNAT instance. (For the purposes of this article, we use the term ``deidentification'' to mean the superset of ``(complete) anonymisation'' and ``pseudonymisation''.)
    
        Although flexible abilities to filter data are now built into DicomEdit 6, this was not always the case, so our application has an independent facility to reject unwanted DICOM information objects by filtering on the SOPClass UID tag (0008,0016). Historically, the internal research projects undertaken have not involved either secondary capture DICOM or other “higher risk” modalities such as ultrasound. This has enabled us to impose a regime whereby our application filters out such potential sources of “burned-in” protected healthcare information (PHI) data, in an added layer of security prior to the deidentification itself. 
    
        For projects involving external partners (e.g., multicentre trials), workflow is more varied and complex, reflecting individual circumstances of the healthcare organisations generating data. Sites have a responsibility (and sign data sharing agreements to this effect) to ensure data are deidentified on-site prior to being transmitted to us. Our preferred method is for sites to use the XNAT Desktop Client (XDC). As above, the software communicates with our destination XNAT server, retrieves the anonymisation script associated with the relevant trial, and deidentifies the data at source. Where use of XNAT Desktop Client is not possible, collaborating institutions use their own in-house methodologies for anonymisation. We have observed that these can be inflexible and lacking in the sophisticated facilities provided by our preferred route, and, for this reason, we normally add an additional layer of deidentification when data are received by our systems, to rectify where possible issues caused by this (see below).
        
        Prior to transmission of the data, site staff are required to inspect outgoing images visually to screen for the presence of burned-in PHI. At the same time, we add a filter on the SOPClass UID, as before, because our experience shows that sites may download and forward DICOM studies from PACS that inadvertently contain modalities not requested by the trial protocol.

    \subsection{Use cases}
        As part of our work managing the joint ICR and RM XNAT Repository, we deal with a number of different research use cases, and each of these requires a correspondingly different approach to the use of deidentification tools. Our overall approach is to apply {\em attribute confidentiality profiles}, as described in Part~15, Section~E of the DICOM standard, consisting of the Basic Profile modified by a combination of options. For each of these, we aim to provide a full risk assessment, together with an example dataset and validation methodology, in order to create ``unit tests'' that can be run each time a change in the deidentification script is made. The Medical Image Deidentification Benchmark (MIDI-B) challenge\footnote{\url{https://www.synapse.org/Synapse:syn53065760/wiki/625274}}, discussed in detail below, thus exemplifies one of a suite of different profiles that we support, viz:
        \begin{itemize}
            \item {\em maximal anonymisation} via use of the DICOM Basic profile
                
            This might be appropriate for a machine learning application that needs little beyond the image voxel values. Even in this straightforward case, where no patient-related metadata are required, the rigour of the DICOM standard still provides advantages for data-sharing over less metadata-rich formats, because it describes {\em inter alia} how image series link together, for example, the association between regions-of-interest (ROIs), represented by DICOM RT-STRUCT contours or DICOM-SEG masks, and the ``base'' images that these ROIs annotate.

            \item deidentification with {\em maximal retention of scientifically useful information}
                
            This is the MIDI-B use case and represents a considerable challenge because of the need to implement the Part~15~E Clean Descriptors option. This is typically appropriate where data are placed in a public repository such as The Cancer Imaging Archive\footnote{\url{https://www.cancerimagingarchive.net/}}.

            \item deidentification for use in the context of a {\em multicentre clinical trial with patient consent to participate}
                
            Whilst it is still a requirement that subject identity is not revealed to data analysts and other members of the trial team, the list of users with access to the data will be highly restricted. Staff operate under a contract and the risk of certain threats (such as reverse-engineering of DICOM files to attempt reidentification) is correspondingly reduced. It may also be vital for the purposes of study management and avoidance of error that the Retain Longitudinal Temporal Information With Full Dates Option be used (i.e., no date shifting is applied), and this is permitted within our IG framework. Individual clinical trials might also have particular requirements for specific patient demographics, or procedure codes to be retained. Explanation of all these features is included in participant information sheets and informed consent is given.  Finally, submission of the data to the trial sponsor might necessitate custom ``data routing'' information to be inserted into specific DICOM tags to allow the data to be correctly archived by the destination repository.

            \item {\em internal use of data}
                
            For data that are restricted to internal networks of hospitals or their associated academic institutions, certain confidentiality profile options, such as Retain Institution Identity, Retain Device Identity and Retain UIDs may be beneficial from an organisational perspective and may reveal no information that is not already known by the users of the data.

            \item{\em ``repair'' of faulty prior attempts at deidentification}
                
            Correct deidentification, as illustrated by the findings of the MIDI-B challenge, is a complex process that requires an in-depth knowledge of the DICOM standard that is not available at all sites providing data for research studies. In our experience of running a service that receives images from more than 60 partners worldwide, many institutions have inflexible internal procedures for how data should be deidentified, and, unfortunately, in some cases this leads to non-compliant DICOM files. As illustrative examples, we cite the replacement of the original contents of some string fields that have DICOM ``value representation'' (VR) of Long String (LO) with strings of length more than 64 characters, and the replacement of the contents of the Study Date (0008,0020) tag with the value "ANON", which is likely to cause downstream applications to fail, because the DICOM standard requires date values to have a VR of Date (DA). We have also observed the modification and renaming of private vendor blocks, which, while still conforming with the DICOM standard and retaining necessary information, can lead to unexpected downstream effects. 

            \item {\em reconstruction of non-conforming DICOM}

            Closely linked to the above item is the topic of augmenting received DICOM files with imputed values for tags that are missing but required for conformance with the DICOM standard. MIDI-B incentivised participants to ``improve the quality'' of incoming data. A straightforward task here might be the addition of missing Type 2 DICOM tags (i.e., ones that are required to exist by the standard but may be left empty if the value is unknown).  Potentially more controversial is what to do in cases where required tags that {\em should contain real, non-null data} (i.e., Type 1 tags) are missing. For fields included in the Part~15~E Basic Profile, the standard requires a ``a non-zero length value that may be a dummy value and consistent with the VR''. For some information object definitions (IOD), the relevant tag to be replaced might correspond to a complex, nested DICOM \textit{sequence} for which no easy dummy value can be used, and might involve imputation of multiple mandatory fields in a way that is not straightforward.
        \end{itemize}

    \subsection{Rationale for participating in MIDI-B}
        It is imperative that XNAT support effective deidentification processes to prevent inadvertent storage of PHI outside of an appropriate IG context. Several aspects of the processes described above benefit from our participation in the MIDI\nobreakdash-B challenge.
        \begin{itemize}
            \item The challenge allows us to benchmark our existing methods against a varied set of data with known insertions of ``fake'' protected health information (PHI) / patient identifiable data (PID), and have them evaluated independently according to an external scoring system.
            \item Many of the tags in the varied MIDI dataset are ones we had never previously encountered ``in the wild'' in more than 15 years of anonymising data for multicentre studies.
            \item Our primary clinical trial use cases do not publish data on an open-access basis like TCIA, and thus involve a very different set of choices for the retention and deletion of DICOM tags. We were thus keen to see whether our existing system could be translated merely with appropriate changes to the anonymisation script used, or whether more extensive modifications to the pipeline were required.
            \item A common mode of integration of third-party AI tools into clinical PACS is to have the tool return processed data into the clinical workflow as secondary capture DICOM. Increasingly, these processed images are likely to be essential inputs into clinical research studies. This means that our historical strategy of rejecting secondary capture DICOM is likely to become untenable. The challenge allowed us to test a new method for automatically removing burned-in PHI.
            \item Our current workflow for external sites involves the requirement for site staff to inspect images visually. Testing an automated tool that could, in future, be deployed at sites to remove burned-in PHI might lead, ultimately, to a significant reduction in workload for site staff.
        \end{itemize}

\section{Related Works}
    Several toolkits for reading, manipulating and rewriting DICOM files --- the necessary prerequisites for deidentification of medical images --- exist in a variety of high-level computer languages. Commonly used examples are the pydicom\footnote{\url{https://github.com/pydicom/pydicom} and specific deidentification tools in \url{https://github.com/pydicom/deid}} (Python), DCM4CHE\footnote{\url{https://web.dcm4che.org/}} (Java) and DCMTK{\footnote{\url{https://dicom.offis.de/en/dcmtk/dcmtk-software-development/}}} (C/C++) libraries, but this list is not exhaustive. Built on top of such libraries is an ecosystem of more specialist libraries and applications specifically aimed at the deidentification of DICOM data, which has been an area of applied academic study for several decades.

    \cite{rodriguez2010open} described five tools for DICOM deidentification existing at the time, whilst \cite{shahid2022two} summarised recent work, and \cite{kondylakis2024documenting} described the approach to deidentification from five flagship projects from the European Union's AI for health imaging (AI4HI) programme. Popular current tools include DicomCleaner\footnote{\url{https://www.dclunie.com/pixelmed/software/webstart/DicomCleanerUsage.html}} and the \textit{RSNA MIRC Clinical Trials Processor (CTP)}\footnote{\url{https://mircwiki.rsna.org/index.php?title=MIRC_CTP}} \citep{freymann2012image}, which is the basis of the deidentification workflow in the National Biomedical Imaging Archive Project and other institution-specific pipelines (e.g., \citep{mesterhazy2020high}). DicomEdit, as used in this work, was built on top of the DCM4CHE toolkit and, although it can be used in a standalone fashion, has been developed largely for use within XNAT and related applications (such as {\em XNAT Desktop Client}\footnote{\url{https://www.xnat.org/download/}} from the XNAT core development team, and our own {\em anontool}). Both CTP and DicomEdit are script-based solutions, allowing very flexible definition of deidentification specifications. Microsoft provides an open-source anonymisation solution\footnote{\url{https://github.com/microsoft/Tools-for-Health-Data-Anonymization}} with a similar philosophy, allowing anonymisation profiles to be specified in JSON format.

    In addition to these methods of on-premises data deidentification, both AWS HealthImaging\footnote{\url{https://aws.amazon.com/blogs/machine-learning/de-identify-medical-images-with-the-help-of-amazon-comprehend-medical-and-amazon-rekognition/}} and Google Healthcare\footnote{\url{https://cloud.google.com/healthcare-api/docs/how-tos/deidentify}} provide cloud-based solutions.

\section{Methods}
    \subsection{Overview}
\begin{figure}[h]
	   	\centering
		  \includegraphics[width=1.0\linewidth]{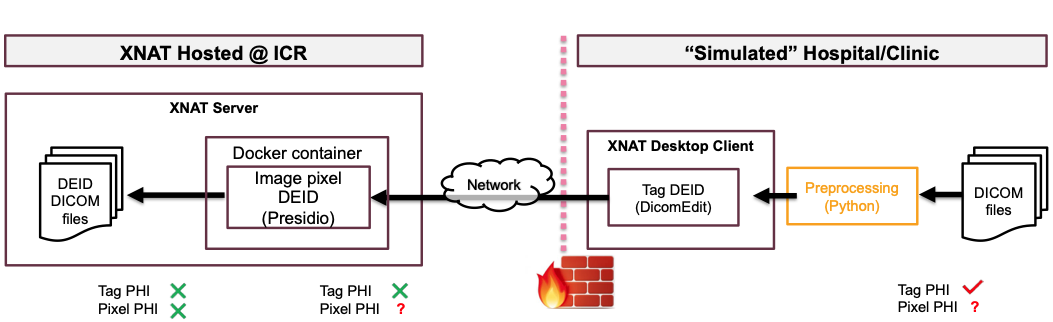}
            \label{fig1}
		\caption{Overview of PHI-removal process for studies imported into XNAT. Note that the Python data preprocessing and the Docker image pixel processing on XNAT are not part of our standard clinical trials workflow, and were added specifically for the MIDI-B challenge. We expect the Python preprocessing step not to be needed in real-world applications, because we can make the reasonable assumption that data direct from clinical systems are DICOM-compliant, whereas several of the MIDI-B manipulated datasets were internally inconsistent and needed ``fixing'' at the start of the pipeline.}
	\end{figure}

	Fig. \ref{fig1} shows an overview of the MIDI-B deidentification process for XNAT, consisting of the following main stages:
        \begin{enumerate}
            \item preprocessing of the DICOM data to enforce harmonisation of tags across series;
            \item deidentification of metadata using either DicomEdit 6 (our standard clinical trials workflow), or Microsoft Presidio\footnote{\url{https://microsoft.github.io/presidio/}} \citep{patchipala2023data,albanese2023text} or a combination of the two; 
            \item uploading of the processed DICOM files to the XNAT server (achieved in real-world situations using either XNAT Desktop Client or XNAT's built-in zip-file uploader);
            \item removal of pixel-based (``burned-in'') PHI via a containerized version of the Microsoft Presidio pixel deidentification software --- this step is currently experimental and, if proven effective could plausibly be incorporated into tools such as XNAT Desktop Client and performed ``client-side''.
        \end{enumerate}
    
    \subsection{PHI removal from DICOM metadata using DicomEdit}
        The DicomEdit~6 Java library accepts an “anonymisation script” consisting of deidentification operations on the source file’s DICOM tags, such as deletion, replacement with a default value, date-shifting and UID “hashing”. Rather than producing streamlined scripts appropriate for particular modalities or local use cases, our strategy for encoding the requirements of Part 3.15 Section E of the DICOM standard has been to follow Table E.1-1 as closely as is possible within the framework of the DicomEdit language. Indeed, many of the recent developments in the library have been driven by this process. Our goal was, thus, to produce a set of scripts that are universally applicable for all DICOM IODs.
    
        DICOM’s design includes subtleties that make rigorous adherence to the standard challenging. For some tags, Table E.1-1 lists the recommended action for the Basic Confidentiality profile as X/Z/D. This means that under some circumstances, the tag should be removed altogether, whilst under others, the existing contents should be replaced by either a zero-length or non-zero length dummy value. The required behaviour is IOD-dependent and, for conditionally required tags (type 1C), might be dependent on the contents of other tags within the same DICOM object. A known weakness of our previous processes was the handling of the “D” action (“replace with a non-zero length value that may be a dummy value and consistent with the VR”) as applied to DICOM sequences. A particular case in point is the Type 3 (optional) tag (0040,A730). This is a complex, potentially recursively-nested sequence, with multiple conditionally required fields. The tag has previously not been present in the types of DICOM file we receive into our imaging studies, and so had not been studied by us prior to the challenge. Understanding what the “correct” default value should be in such cases is important. A plausible approach here is to start with the sequence in the file to be deidentified (which the standard recommends should be entirely replaced by the dummy value) and, instead, recursively anonymise each sub-sequence using the standard rules. At the lowest level, the bottom ``leaf'' tags can be replaced with simple dummy defaults that can be ``hard-coded''. Arguably, this is preferable to strict adherence to the standard, as values in subsequences that do not require anonymisation are not replaced by a dummy, thus potentially retaining valuable information.
    
        Via manual review of Table E.1-1 of the DICOM standard — our original work was based on version 2023b — we developed a script providing a “best-efforts” implementation of the Basic Confidentiality profile, considering each tag in turn, determining possible complications, and taking pragmatic decisions. We then automated the process of applying the various “retain” options defined by Part~3.15 Section~E to allow the creation of a range of standards-compliant scripts employing different combinations of options. Finally, we constructed a script that implemented the options needed to match the requirements of the MIDI-B challenge, and this was used to deidentify the DICOM files to be uploaded to the Synapse server for the challenge.

    \subsection{Implementation of the ``Clean Descriptors'' option}
        The most challenging difference between our routine clinical trials scenario and that tested by the MIDI-B challenge was the requirement to implement the Clean Descriptors option of the DICOM standard. For our clinical trials work, we are normally able to remove completely all free-text fields, because the relevant metadata are available from other clinical information independently gathered as part of the trial. The exception is tag (0008,103E) Series Description, which is of vital importance in MRI but one for which we have not historically encountered issues of embedded PHI, despite it being a field that can be edited by technicians at the scanner. MIDI-B deliberately introduces PHI into this field.

        Our strategy for the MIDI-B challenge was pragmatic, with the following steps:
        \begin{enumerate}
            \item Simple ``rules-based'' approach

            The aim here was to decide whether a full machine-learning solution was, in fact, necessary. We used DicomEdit to implement regular expressions that removed:
            \begin{itemize}
                \item any text matching the words in the Patient Name (0010,0010) or Patient ID (0010,0020) tags
                \item all text following ``trigger words'' commonly found associated with names of patients, medical staff or hospitals/clinics in the MIDI-B validation data (specifically ``for'', ``by'', ``at'', ``to'', ``on'');
                \item date strings of form \verb|yyyymmdd| and any occurrence of a year in the 20th or 21st Century;
                \item common patterns of strings of digits and punctuation characters that might be indicative of phone, credit card or social security numbers, or patient IDs (specifically, groups of 9 or more consecutive characters all from the set \{digits, parentheses, dashes and the letter x\}).
            \end{itemize}

            \item Microsoft Presidio NLP using imaging metadata

            To compare against the relatively simple DicomEdit approach, we used Microsoft's Presidio framework to apply several exemplar Natural Language Processing (NLP) Named Entity Recognition (NER) approaches as "recognizers". In the first experiment, we applied consecutively the following two models:
            \begin{itemize}
                \item  FLAIR\footnote{Amazon machine learning blog \textit{ibid}} \citep{FLAIR} for identification of "non-medical" text such as locations and addresses;
                \item StanfordAIMI-deidentifier-base\footnote{\url{https://github.com/MIDRC/Stanford_Penn_MIDRC_Deidentifier}} \citep{StanfordAIMI} for identification of medical terminology with a specific focus on the radiology domain;
            \end{itemize}
 
            whilst in the second, we used a version of a RoBERTa model \citep{liu2019roberta} fine-tuned for deidentification of medical notes\footnote{\url{https://huggingface.co/obi/deid_roberta_i2b2}}. This was intended to be a brief scoping survey to investigate the potential of such methods, with the aim of discovering the types of issues that might arise, rather than an exhaustive search for the best available algorithm.
            
        \end{enumerate}

    \subsection{Removal of PHI ``burned into'' image pixels using XNAT Desktop Client and Microsoft Presidio}
        The XNAT Desktop Client provides a number of advanced tools allowing human data uploaders to streamline the process of manual removal of burned-in PHI. These include the ability to define pixel masks, and a semiautomatic series-grouping tool to facilitate application of these masks to multiple series. Nevertheless, in our current practice, there remains no easy way of avoiding the need for a real person to inspect every image, however briefly.   
        
        For our experimental implementation, pixel-level deidentification is carried out in XNAT itself. Incoming files are temporarily stored in a protected “prearchive”, to which we do not allow user access. New uploads trigger an event that causes a Docker container to run a process based around the Microsoft Presidio library. Here, the goal of Presidio is to de-identify any text found within images by an optical character recognition (OCR) process. Presidio supports more sophisticated analysis models via cloud-based Azure services, but our local IG mandates only “on-site” processing at this time. Hence, the default PHI text library \citep{spaCy} and image OCR library \citep{Tesseract, smith2007overview} were used, with the \verb|use_metadata| option set to true to augment the text analyzer with key DICOM tag information such as (0010,0010) Patient Name. Outputs of the container process are DICOM files with redaction boxes masking out any identified PHI. 

\section{Results}
	\subsection{Result of original MIDI-B challenge}
            Our initial submission in the validation phase, using a largely unmodified script from one of our local use cases — with DICOM options Retain Safe Private (our normal clinical trial list of tags), Retain Patient Characteristics and Retain Longitudinal Modified Dates — scored 92.8\%. Analysis showed two main classes of discrepancy:
            \begin{itemize}
                \item massive under-retention of private tags that TCIA considers to be safe: our clinical trials policy is to remove private tags not explicitly shown to be important for studies;
                \item lack of attempt to extract useful information from fields to be cleaned: our prior internal algorithms have explicitly not attempted to implement the Clean Descriptors option, but instead used the basic profile for these fields.
            \end{itemize}

            \begin{table*}[t]
		      \centering
		      \includegraphics[width=0.95\linewidth]{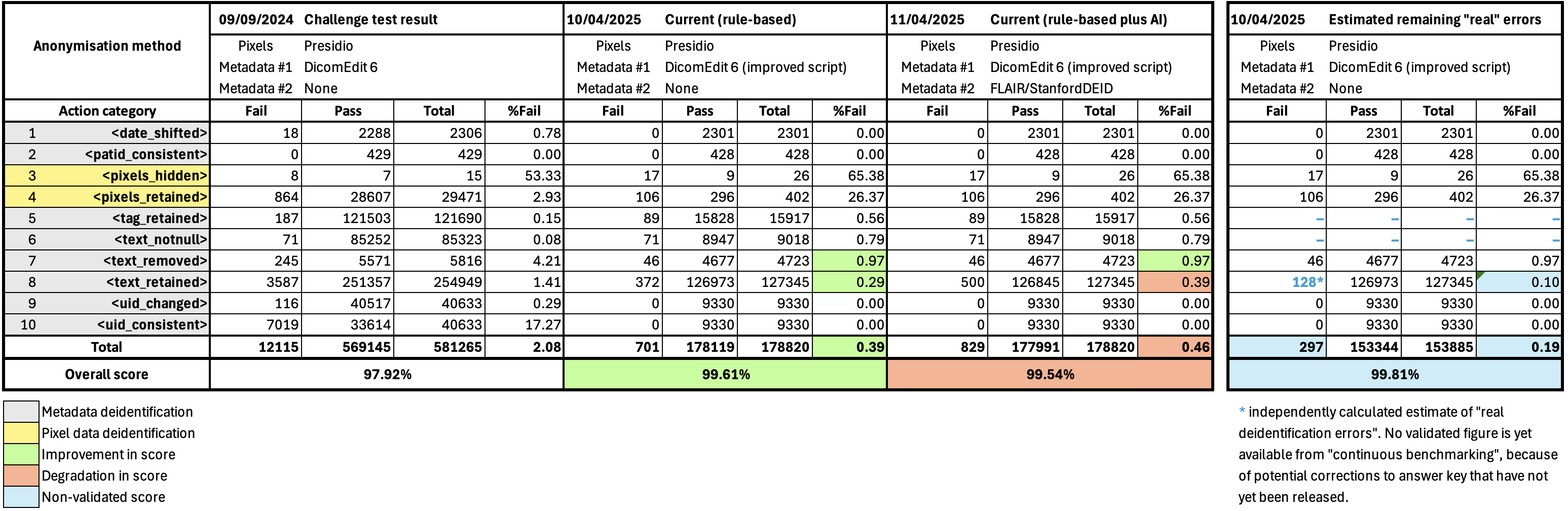}
                \caption{Results from the test set of the original MIDI-B challenge (September 2024) and our subsequent refinements processed via the MIDI-B Continuous Benchmarking facility. The right-hand block shows our estimate of residual errors after issues currently under discussion with the challenge organisers (and not yet included in the benchmarking ``answer key'') are taken into account.  Categories of deidentification ``error'' are shown in the first column, where the ``action'' relates to the operation that {\em should} have occurred. Total columns do not match between the challenge and later benchmarking runs, because the basis of the scoring was updated by the organisers; nevertheless, percentages failure percentages should be broadly comparable. Notes: (1) Simple script error where tag from Table E.1-1 was inadvertently omitted for scoring run, easily rectified; (3,4) See the Discussion regarding the performance of Microsoft Presidio for ``burned-in'' PHI. (5,6) We are not convinced that the goal of deidentification should be to “invent” data for non-existent tags to make files more DICOM-compliant, and so did not attempt to reduce these ``failures'' to zero; (7,8) See Discussion; (9) Trivial single character or one-line mistakes in the anonymisation script were corrected at our next iteration of the anonymisation script; (10) This was a bug in our UID mapping file generation for the challenge and the DICOM files themselves were anonymised correctly.}
                \label{series_level_scores}
            \end{table*}
        
            Updating our process to account for these issues led to two challenges:
            \begin{enumerate}
                \item We observed that many of the series artificially created for the challenge had PHI inserted into the metadata only for a subset of SOPInstances. For example, the Series Description tag was modified to include patient names only for a few slices of some image series. Whilst each DICOM file was individually valid, there was a serious data consistency problem at a series level, with multiple values of certain key tags across SOPInstances with the same SeriesInstanceUID. In our reading of the DICOM standard we have found nothing that appears to prohibit this, yet to all intents and purposes, these are “broken” DICOM series. XNAT (rightly, in our view) refuses to archive such data and other image viewing tools either refuse to display the original series or, instead, split the single series into multiple separate items for display, based on the unique values of the Series Description or Series Number.
                
                Our response was to insert the preprocessing step of Fig. 1, replacing  the relevant tags corrupted with PHI and ensuring that all series were internally consistent. This resulted in a reduction in the number of instance-level warnings and errors of almost 400k, equating to an 84\% improvement (49,897 errors in baseline to 6,956 in our submission). This fix was easy to apply, but the downside was that it could not be handled using DicomEdit, because that operates independently on each SOPInstance. Hence, an extra preprocessing step was required.
                
                \item More problematically, when we included TCIA’s more extensive set of “safe” private tags, we discovered a previously unreported issue with the output of DicomEdit that occurs when DICOM files encoded using the implicit value representation (VR) transfer syntax are processed with DicomEdit's \verb|retainPrivateTags| directive. Analysis of the open-source code showed the order of operations in the algorithm to be: (1) compile a list of private tag values for the tags to be retained; (2) delete all private tags; (3) rewrite the retained private tags to the output DICOM file. For files with implicit VR, the “correct” VR is by definition unknown; indeed, in the TCIA-supplied .csv file listing the “safe” tags to keep, a large number of tags have multiple associated VRs.
                
                This leads to the unfortunate circumstance in which DicomEdit makes a different assumption about the output VR for some tags from the one pydicom uses to read the data. An error condition then occurs when a Python application attempts to read the rewritten DICOM file. Further compounding the problem, this mismatch caused the TCIA assessment script on the Synapse competition platform to fail altogether with our processed files, meaning that no discrepancy report was generated, and this meant we were not able to conduct further validation runs, compromising our anonymisation script development.
                
                We eventually solved the problem prior to the final test by changing the behaviour of our private fork of the DicomEdit repo to: (1) compile a list of private tags present in the file but not in the retain list; (2) delete these tags. Since this operation does not involve rewriting tags, pydicom can read the resulting file. Correction of this issue will be put in place by the core XNAT development team in a future release of DicomEdit and is contingent of an upgrade of the entire library to the latest version of the underlying DCM4CHE toolkit.
            \end{enumerate}

            Our final challenge score was 97.91\%, and, in Table~\ref{series_level_scores}, we show the test score broken down according to the TCIA “action” categories, together with the types of error and the necessary mitigations. These results were disappointing, but largely reflected the fact that (as a result of the technical difficulties above) several trivial mistakes had not been caught prior to the test submission. 

        \subsection{Additional work following the challenge}
            By the time of the MIDI-B workshop in October 2024, the initial results had been analysed and corrections to our scripts had been incorporated, leading to a much improved score of 99.54\%. Further work since that time has been focused on reducing errors in the \verb|<text_removed>| and \verb|<text_retained>| categories via continued refinement of our DicomEdit script, the investigation of two large language models, and an ongoing discussion of remaining categories of error with the challenge organisers.

            The second block of data in Table~\ref{series_level_scores} shows that a purely rules-based approach using DicomEdit alone for metadata anonymisation reduces errors in the \verb|<text_removed>| category to 46, close to a 1\% failure rate, with \verb|<text_retained>| errors reduced to 372, a failure rate of just 0.29\%.

            The third block of data in Table~\ref{series_level_scores} shows the result of applying the combination FLAIR+StanfordDEID model after our anonymisation using DicomEdit. The \verb|<text_removed>| category surprisingly remained at 46 (although with a slightly differing distribution of errors), while the number of series errors in the \verb|<text_retained>| category rose from 372 to 500. In our second LLM experiment, the clinical RoBERTa model {\em was} able to further reduce to 28 the number of \verb|<text_removed>| errors, but, again, at the cost of increasing the number of \verb|<text_retained>| errors, this time to between 498 and 627, depending on the number of tags that the model was instructed to inspect.

            Examples of the types of error for DicomEdit alone, and DicomEdit followed by FLAIR+StanfordDEID and clinical RoBERTa are presented in Tables S1, S2 and S3 respectively of the supplementary data. 

            The fourth block of data represents our internal estimate --- i.e., not officially ratified results from the MIDI-B Continuous Benchmarking facility --- of errors that represent genuine failures of our deidentification process. These figures reflect post-challenge technical discussions with the organisers, that had not yet been incorporated into the official answer key at the time of manuscript submission.

            Completely outside the challenge context, but still relevant here, we have also shown in an independent project that the same methodology is able to deidentify DICOM-SEG region-of-interest data (an IOD not included in the MIDI-B corpus) and maintain its linkage with the related base MR images, whilst still replacing the DICOM UIDs (i.e., the Basic Application Level Confidentiality Profile was not modified to include the Retain UIDs option). 

    \section{Discussion}
        We classified the issues raised by the MIDI-B challenge into a number of broad categories:
        \begin{itemize}
            \item {\em Genuine mistakes in our scripts and issues with the DicomEdit code library}
            
            These were swiftly corrected and are not a problem for future processing.

             \item {\em PHI that needs to be removed}
        
            Of most concern for us is the MIDI-B error category \verb|<text_removed>|. Errors here correspond to PHI that should have been removed from the DICOM metadata but was retained erroneously by our deidentification process, according to the answer key provided by the challenge organisers. Success relates directly to our ability to properly implement the DICOM Clean Descriptors confidentiality profile option. Historically, our policy has been conservative, and we have previously chosen to remove fields for security, rather than risk deidentification failures. MIDI-B penalised this approach and, instead, incentivised maximum retention of potentially valuable scientific information. For the MIDI-B task, we achieved excellent performance via simple heuristics: (i) searching for “trigger” prepositions (``in'', ``for'', ``by'', ``at'' and ``on'') and deleting text after these; (ii) applying regular expressions to remove groups of numbers and punctuation characters that might represent phone, social security, accession number or bank details. However, this approach is inherently “brittle”, and whilst the methodology can easily be extended on a case-by-case basis to data from non-anglophone countries, we are concerned that performance on the MIDI-B dataset would not generalise to a truly mixed international dataset. Nevertheless, an additional safeguard is implemented by removing all occurrences of text found in the Patient Name (0010,0010) and Patient ID (0010,0020) DICOM tags, wherever these occur in other free-text fields in the DICOM file.
            
            For this challenge, we demonstrated that our rules-based approach was able to detect and remove in excess of 99\% of all PHI that should have been removed (aggregated at a series level). Out of the whole dataset (432k \verb|<text_removed>| actions to take), only two actual instances of patient names were missed. These occurred for very deeply nested instances of a widely used tag (0040,A160), which represents a generic ``Text Value'' part of a Content Sequence attribute. (This type of tag is typically found in structured reports, but {\em does not} appear to be separately listed in Table~E.1-1 of the DICOM standard.) DicomEdit currently lacks the facility to perform the optimal redaction of this type of sequence: the names could be removed, but this would come at the cost of increased penalties in the \verb|<text_retained>| category. We plan to log a feature request with the XNAT development team.
            
            Almost all the other \verb|<text_removed>| errors related to the inability of our heuristics to detect {\em addresses} reliably. This represents a HIPAA violation and could have resulted in the identities of 21 patients from the test set being compromised, an unacceptable total, hence the need to investigate additional methods.

\begin{table*}[t]
    \begin{center}
        \begin{tabular}{|c|l|}
            \hline
            Original & Mark Wilcox : 908 E Maryland Ln Laurel, MT 59044\\
            \hline
            Masked & \_PNB\_ \_PNA\_ : 908 E {\bf LOCATION} Ln {\bf LOCATION}, {\bf LOCATION} 59044\\
            \hline
        \end{tabular}
    \end{center}
    \caption{Example of partial removal of an address element from the (0010,21B0) "Additional Patient History" DICOM tag from one of the challenge datasets. \_PNA\_ and \_PNB\_ represent name tokens successfully masked via DicomEdit, whilst each {\bf LOCATION} entry represents a partial address token detected by Presidio in conjunction with the FLAIR+StanfordDEID model combination as ``recognizers''.}
    \label{tab:location_results_compare}
\end{table*}

            Our initial work to reduce address-related failures explored the application of Microsoft Presidio (via the wrapped spaCy toolkit combined with three different transformer models: FLAIR and StanfordAIMI-deidentifier-base used consecutively, and a RoBERTa model fine-tuned for clinical vocabulary). For Experiment~1, the ML model combination made no difference to the \verb|<text_removed>| result under the scoring algorithm used by MIDI, as typically the ML only partially redacted elements of addresses when complete removal of all elements was required. As an example, consider the instance of (0010,21B0) "Additional Patient History" tag shown in Table~\ref{tab:location_results_compare}. The patient name was removed by DicomEdit (replacements bracketed by underscores), but only some of the location elements were identified by Presidio (replacements in bold capitals). It is likely that a more specific and aggressive address identification method could resolve such issues. The algorithm of Experiment~2 was slightly more successful, but overall, both ML model combinations made the overall score worse because of the issues below.

            \item {\em Scientifically useful text that should be retained}

            The ML models studied had a markedly negative effect in the \verb|<text_retained>| category and were found to be inaccurate and too aggressive, frequently mis-identifying parts of the text that should be retained for medical utility at the level of preservation required by MIDI-B. Additionally, it was found that --- in contrast to the predictability of the DicomEdit outcomes --- model choice and small changes in the ML algorithm parameters had large effects on the results leading to concerns as to the reliability and over-sensitivity of such approaches, unless significant work has gone into ensuring the robustness and ongoing real-world performance. In particular, it is extremely difficult to determine why a given model has made a specific decision, making root-cause analysis of errors challenging.
            
            Other errors \verb|<text_retained>| errors have been the subject of on-going technical discussions with colleagues from the MIDI-B organising team. 

            \item {\em Complexities of the DICOM standard}

            We are concerned that IOD-dependent behaviours, related to where Table E.1-1 of Part 3.15 of the DICOM standard specifies an action of form ``X/Z/D'' may lead to excessive complexity in anonymisation scripts if rigorous adherence to the standard is required.  
            
            \item {\em Metadata changes to improve conformance of data to the DICOM standard}

            As part of the challenge, we encountered occurrences of inconsistencies in datasets or non-conformities with the DICOM standard that would either have prevented us archiving the data at all or would have caused downstream failures of software (for example, standard imaging software failed to display the data correctly). The marking scheme of the MIDI-B challenge explicitly incentivised participants to ``repair'' such data. However, we question whether this is properly the role of ``deidentification''. In our view, a balance needs to be struck between enabling our users to obtain maximum scientific value from ``broken'' data, and feeding back to providers (forcibly, if necessary, by rejecting their data as non-compliant) that they need to improve the quality of image metadata at source.   
             
            \item {\em Private tags}

            By definition, private tags are not required to conform to a published standard. Vendors can and do change the contents of these fields with successive scanner software releases in ways that may cause downstream processing to fail if it makes assumptions based on previous versions. The MIDI-B challenge used a compilation\footnote{\url{https://wiki.cancerimagingarchive.net/download/attachments/3539047/TCIAPrivateTagKB-02-01-2024-formatted.csv?version=2&modificationDate=1707174689263&api=v2}} of the imaging community's current state of knowledge of the content of manufacturer private tags, created by staff at the The Cancer Imaging Archive, as the gold standard for deciding how to handle these tags. In our view, it remains an open question as to whether one can rely on this type of information to drive a fully automated deidentification process to the extent that human checks are not required.

            \item {\em ``Burned-in'' PHI}

            Our estimated final performance of 99.81\% represents an upper ceiling for this methodology, because of our remaining failures in removing ``burned-in'' PHI from some challenge images. This area of work was new to us at the start of the challenge, and we confined our research to the evaluation of a readily-available existing tool, Microsoft Presidio. There were relatively few examples of such images in the MIDI-B corpus and it is hard to simulate the full range of different presentations of burned-in data, particularly when considering both historical data (e.g., scanned plane-film mammograms with variable annotation characteristics) and unfamiliar new forms of data (e.g., microscope slides, with either hand-written or barcode labels). Medical image text detection also presents particular difficulties related to low-resolution images and partially occluded or low-contrast text. Out-of-the-box, Tesseract may not have been trained against this type of target and, thus, we suspect that current state-of-the-art solutions, coupled with suitable additional training data, could substantially improve on our results here.
        \end{itemize}

    \section{Conclusion}
        We have demonstrated a highly successful methodology for deidentifying DICOM data within the XNAT ecosystem, validated as part of the MIDI-B challenge, and have outlined a number of different use cases for which our framework is in current use. We estimate our overall performance to be 99.81\%.
        
        A purely rule-based (i.e., non machine-learning) solution was shown to be capable of complete redaction of all patient, staff and hospital name information from the MIDI-B test corpus, but had an overall 1\% error rate in PHI removal, with failures occurring almost exclusively for {\em address-}related PHI. It was shown that this figure could be improved by incorporating machine-learning, but at a significant cost of over-redaction of non-PHI text from other fields.  A number of outstanding issues still remain, requiring future research, the most pressing of which is improved removal of patient-identifying information ``burned into'' image pixels. An important factor limiting future development is the need for a larger ``real-world'' dataset of burned-in PHI for further model training and testing.

%


\acks{This project represents independent research funded by the National Institute for Health and Care Research (NIHR) Biomedical Research Centre at The Royal Marsden and Institute of Cancer Research, the NIHR Royal Marsden Clinical Research Facility and by the Royal Marsden Cancer Charity. AM is part-funded by Cancer Research UK grant RRNIA{\textbackslash}100001, and infrastructure used for this project was developed the National Cancer Imaging Translational Accelerator, also funded by Cancer Research UK. The views expressed are those of the authors and not necessarily those of the NIHR, the Department of Health and Social Care or the other organisations above.}

%
\ethics{The work here used data curated and prepared specially for the MIDI-B challenge, as part of the programme described in \citep{rutherford2021dicom} .}

\coi{There are no conflicts of interest to declare.}

\data{All source data used in this work were downloaded as part of the MIDI-B challenge from the Synapse website\footnote{\url{https://www.synapse.org/Synapse:syn53065760/wiki/625274}}}. Supplementary Data (Tables S1, S2 and S3) are supplied in an Excel spreadsheet accompanying the main paper.

\bibliography{MIDI-B}

\begin{thebibliography}{17}
\providecommand{\natexlab}[1]{#1}
\providecommand{\url}[1]{\texttt{#1}}
\expandafter\ifx\csname urlstyle\endcsname\relax
  \providecommand{\doi}[1]{doi: #1}\else
  \providecommand{\doi}{doi: \begingroup \urlstyle{rm}\Url}\fi

\bibitem[Akbik et~al.(2019)Akbik, Bergmann, Blythe, Rasul, Schweter, and Vollgraf]{FLAIR}
Alan Akbik, Tanja Bergmann, Duncan Blythe, Kashif Rasul, Stefan Schweter, and Roland Vollgraf.
\newblock {FLAIR}: An easy-to-use framework for state-of-the-art {NLP}.
\newblock In \emph{{NAACL} 2019, 2019 Annual Conference of the North American Chapter of the Association for Computational Linguistics (Demonstrations)}, pages 54--59, 2019.

\bibitem[Albanese et~al.(2023)Albanese, Ciolek, and D'Ippolito]{albanese2023text}
Federico Albanese, Daniel Ciolek, and Nicolas D'Ippolito.
\newblock Text sanitization beyond specific domains: Zero-shot redaction \& substitution with large language models.
\newblock \emph{arXiv preprint arXiv:2311.10785}, 2023.

\bibitem[Bauermeister et~al.(2020)Bauermeister, Orton, Thompson, Barker, Bauermeister, Ben-Shlomo, Brayne, Burn, Campbell, Calvin, et~al.]{bauermeister2020dementias}
Sarah Bauermeister, Christopher Orton, Simon Thompson, Roger~A Barker, Joshua~R Bauermeister, Yoav Ben-Shlomo, Carol Brayne, David Burn, Archie Campbell, Catherine Calvin, et~al.
\newblock The dementias platform uk (dpuk) data portal.
\newblock \emph{European journal of epidemiology}, 35:\penalty0 601--611, 2020.

\bibitem[Chambon et~al.(2022)Chambon, Wu, Steinkamp, Adleberg, Cook, and Langlotz]{StanfordAIMI}
Pierre~J Chambon, Christopher Wu, Jackson~M Steinkamp, Jason Adleberg, Tessa~S Cook, and Curtis~P Langlotz.
\newblock {Automated deidentification of radiology reports combining transformer and “hide in plain sight” rule-based methods}.
\newblock \emph{Journal of the American Medical Informatics Association}, 11 2022.
\newblock ISSN 1527-974X.
\newblock \doi{10.1093/jamia/ocac219}.
\newblock URL \url{https://doi.org/10.1093/jamia/ocac219}.
\newblock ocac219.

\bibitem[Freymann et~al.(2012)Freymann, Kirby, Perry, Clunie, and Jaffe]{freymann2012image}
John~B Freymann, Justin~S Kirby, John~H Perry, David~A Clunie, and C~Carl Jaffe.
\newblock Image data sharing for biomedical research—meeting hipaa requirements for de-identification.
\newblock \emph{Journal of digital imaging}, 25\penalty0 (1):\penalty0 14--24, 2012.

\bibitem[Gurney et~al.(2017)Gurney, Olsen, Flavin, Ramaratnam, Archie, Ransford, Herrick, Wallace, Cline, Horton, et~al.]{gurney2017washington}
Jenny Gurney, Timothy Olsen, John Flavin, Mohana Ramaratnam, Kevin Archie, James Ransford, Rick Herrick, Lauren Wallace, Jeanette Cline, Will Horton, et~al.
\newblock The washington university central neuroimaging data archive.
\newblock \emph{Neuroimage}, 144:\penalty0 287--293, 2017.

\bibitem[Kondylakis et~al.(2024)Kondylakis, Catalan, Alabart, Barelle, Bizopoulos, Bobowicz, Bona, Fotiadis, Garcia, Gomez, et~al.]{kondylakis2024documenting}
Haridimos Kondylakis, Rocio Catalan, Sara~Martinez Alabart, Caroline Barelle, Paschalis Bizopoulos, Maciej Bobowicz, Jonathan Bona, Dimitrios~I Fotiadis, Teresa Garcia, Ignacio Gomez, et~al.
\newblock Documenting the de-identification process of clinical and imaging data for ai for health imaging projects.
\newblock \emph{Insights into Imaging}, 15\penalty0 (1):\penalty0 130, 2024.

\bibitem[Liu et~al.(2019)Liu, Ott, Goyal, Du, Joshi, Chen, Levy, Lewis, Zettlemoyer, and Stoyanov]{liu2019roberta}
Yinhan Liu, Myle Ott, Naman Goyal, Jingfei Du, Mandar Joshi, Danqi Chen, Omer Levy, Mike Lewis, Luke Zettlemoyer, and Veselin Stoyanov.
\newblock Roberta: A robustly optimized bert pretraining approach.
\newblock \emph{arXiv preprint arXiv:1907.11692}, 2019.

\bibitem[McAteer et~al.(2021)McAteer, O’Connor, Koh, Leung, Doran, Jauregui-Osoro, Muirhead, Brew-Graves, Plummer, Sala, et~al.]{mcateer2021introduction}
MA~McAteer, James~PB O’Connor, DM~Koh, HY~Leung, SJ~Doran, M~Jauregui-Osoro, N~Muirhead, C~Brew-Graves, ER~Plummer, E~Sala, et~al.
\newblock Introduction to the national cancer imaging translational accelerator (ncita): a uk-wide infrastructure for multicentre clinical translation of cancer imaging biomarkers.
\newblock \emph{British journal of cancer}, 125\penalty0 (11):\penalty0 1462--1465, 2021.

\bibitem[Mesterhazy et~al.(2020)Mesterhazy, Olson, and Datta]{mesterhazy2020high}
Joseph Mesterhazy, Garrick Olson, and Somalee Datta.
\newblock High performance on-demand de-identification of a petabyte-scale medical imaging data lake.
\newblock \emph{arXiv preprint arXiv:2008.01827}, 2020.

\bibitem[Patchipala(2023)]{patchipala2023data}
S~Patchipala.
\newblock Data anonymization in ai and ml engineering: Balancing privacy and model performance using presidio.
\newblock \emph{Iconic Research and Engineering Journals}, 6\penalty0 (10), 2023.

\bibitem[Rodr{\'\i}guez~Gonz{\'a}lez et~al.(2010)Rodr{\'\i}guez~Gonz{\'a}lez, Carpenter, van Hemert, and Wardlaw]{rodriguez2010open}
David Rodr{\'\i}guez~Gonz{\'a}lez, Trevor Carpenter, Jano~I van Hemert, and Joanna Wardlaw.
\newblock An open source toolkit for medical imaging de-identification.
\newblock \emph{European radiology}, 20:\penalty0 1896--1904, 2010.

\bibitem[Rutherford et~al.(2021)Rutherford, Mun, Levine, Bennett, Smith, Farmer, Jarosz, Wagner, Freyman, Blake, et~al.]{rutherford2021dicom}
Michael Rutherford, Seong~K Mun, Betty Levine, William Bennett, Kirk Smith, Phil Farmer, Quasar Jarosz, Ulrike Wagner, John Freyman, Geri Blake, et~al.
\newblock A dicom dataset for evaluation of medical image de-identification.
\newblock \emph{Scientific Data}, 8\penalty0 (1):\penalty0 183, 2021.

\bibitem[Shahid et~al.(2022)Shahid, Bazargani, Banahan, Mac~Namee, Kechadi, Treacy, Regan, and MacMahon]{shahid2022two}
Arsalan Shahid, Mehran~H Bazargani, Paul Banahan, Brian Mac~Namee, Tahar Kechadi, Ceara Treacy, Gilbert Regan, and Peter MacMahon.
\newblock A two-stage de-identification process for privacy-preserving medical image analysis.
\newblock In \emph{Healthcare}, volume~10, page 755. MDPI, 2022.

\bibitem[Smith(2007)]{smith2007overview}
Ray Smith.
\newblock An overview of the tesseract ocr engine.
\newblock In \emph{Ninth international conference on document analysis and recognition (ICDAR 2007)}, volume~2, pages 629--633. IEEE, 2007.

\bibitem[spacy.io()]{spaCy}
spacy.io.
\newblock Industrial-strength natural language processing in python.
\newblock URL \url{https://spacy.io}.
\newblock Accessed on 21 March 2025.

\bibitem[Tesseract()]{Tesseract}
Tesseract.
\newblock Tesseract documentation.
\newblock URL \url{https://tesseract-ocr.github.io/}.
\newblock Accessed on 21 March 2025.

\end{thebibliography}


\end{document}